\documentclass[conference]{IEEEtran}
\IEEEoverridecommandlockouts
\usepackage{cite}
\usepackage{amsmath,amssymb,amsfonts}
\usepackage{graphicx}
\usepackage{textcomp}
\usepackage{xcolor}
\usepackage{algorithm}
\usepackage[noend]{algpseudocode}
\def\BibTeX{{\rm B\kern-.05em{\sc i\kern-.025em b}\kern-.08em
    T\kern-.1667em\lower.7ex\hbox{E}\kern-.125emX}}
    
\makeatletter
\def\algbackskip{\hskip-\ALG@thistlm}
\makeatother

\usepackage{amsmath}
\usepackage{caption}
\usepackage{subcaption}
\usepackage{array}
\newcolumntype{P}[1]{>{\centering\arraybackslash}p{#1}}

\DeclareMathOperator*{\minimize}{minimize}

\begin{document}

\title{Adversarial Targeted Forgetting in Regularization and Generative Based Continual Learning Models \\
}

\author{\IEEEauthorblockN{Muhammad Umer}
\IEEEauthorblockA{\textit{Department of Electrical \& Computer Engineering} \\
\textit{Rowan University}\\
Glassboro, USA \\
umerm5@rowan.edu}
\and
\IEEEauthorblockN{Robi Polikar}
\IEEEauthorblockA{\textit{Department of Electrical \& Computer Engineering} \\
\textit{Rowan University}\\
Glassboro, USA \\
polikar@rowan.edu}

}

\maketitle

\begin{abstract}
Continual (or ``incremental'') learning approaches are employed when  additional knowledge or tasks need to be learned from subsequent batches or from streaming data. However these approaches are typically \textit{adversary agnostic}, i.e., they do not consider the possibility of a malicious attack. In our prior work, we explored the vulnerabilities of Elastic Weight Consolidation (EWC) to the \textit{perceptible} misinformation. We now explore the vulnerabilities of other regularization-based as well as generative replay-based continual learning algorithms, and also extend the attack to \textit{imperceptible} misinformation. We show that an intelligent adversary can take advantage of a continual learning algorithm's capabilities of retaining existing knowledge over time, and force it to learn and retain deliberately introduced \textit{misinformation}. To demonstrate this vulnerability, we inject backdoor attack samples into the training data. These attack samples constitute the misinformation, allowing the attacker to capture control of the model at test time. We evaluate the extent of this vulnerability on both rotated and split benchmark variants of the MNIST dataset under two important \textit{domain} and \textit{class} incremental learning scenarios. We show that the adversary can create a ``false memory'' about \textit{any} task by inserting carefully-designed backdoor samples to the test instances of that task thereby controlling the amount of forgetting of any task of its choosing. Perhaps most importantly, we show this vulnerability to be very acute and damaging: the model memory can be easily compromised with the addition of backdoor samples into as little as 1\% of the training data, even when the misinformation is \textit{imperceptible} to human eye.

\end{abstract}


%
\IEEEpeerreviewmaketitle

\section{Introduction}
False memory formation -- the phenomenon where one's memory can be easily distorted through post-event misinformation -- is a common problem in human memory\cite{doi:10.1080/09658211.2019.1611862}. Such misinformation is sometimes self-imposed: a person assures him/herself that they had an experience that in fact did not occur. However, false memory formation can also be implanted by external agents: a malicious entity may provide deliberate and persistent misinformation over a period of time to convince an otherwise unsuspecting victim of the adversary's preferred -- but inaccurate -- version of events \cite{vitelli2012implanting}.

We explore the vulnerability of artificial neural networks (ANNs) to adversarial false memory formation in the context of continual learning of a sequence of tasks. Continual learning (CL) algorithms are used when there is additional knowledge or tasks to be learned from additional data batches or streaming data. Such algorithms are particularly useful when the data distributions change over time, i.e., when the all-too-common assumption of i.i.d. data does not hold. Hence, continual learning algorithms have an important place in machine learning due to many -- and increasing -- number of applications that require such algorithms.  

Continual (or incremental) learning is a challenging problem due to \textit{catastrophic forgetting} \cite{mccloskey1989catastrophic}: the phenomenon where the accuracy of the model on some previously-learned -- but still relevant -- task is partially or completely lost (forgotten) while training to acquire new knowledge for a subsequent task. 
Catastrophic forgetting is characterized by the \textit{stability and plasticity dilemma} \cite{grossberg1988nonlinear}, referring to the difficulty in preservation of past knowledge, while acquiring new knowledge.

We consider two important and practical scenarios of continual learning, namely \textit{domain incremental learning} and \textit{class incremental learning} \cite{vandeven2019three}. Domain incremental learning represents a setting where only the input or marginal data distributions differ between tasks, with class distributions remaining fixed. In contrast, class incremental learning represents a more challenging scenario, where the distributions of both the input marginal data and that of the classes differ between tasks.

The primary question we explore is whether an adversary can take advantage of CL algorithm's ability to retain prior knowledge while acquiring new knowledge, and insert misinformation during training in order to distort the algorithm's memory and deliberately force loss of its prior knowledge.

In our prior work, we showed the vulnerability of a specific regularization-based continual learning algorithm, elastic weight consolidation, to adversarial \textit{backdoor} poisoning attacks with visibly obvious backdoor patterns (such as small subimages) that are embedded into the training dataset with a false label of the attacker's choosing \cite{Umer2020TargetedFA}. In this effort, we extend that work to demonstrate the same vulnerability in other state-of-the-art regularization-based and generative replay-based CL algorithms. We show that the very mechanisms used by CL algorithms to retain prior knowledge can be easily exploited by an adversary and used against the model itself. Furthermore, we now show that an adversary only needs to insert a small amount of misinformation -- that is visually \textit{imperceptible} to the human eye -- in order to capture complete control of the model at test time.



\section{Background \& Related Work}
\subsection{Continual Learning}
\textit{Regularization-based approaches} retain prior knowledge by penalizing changes in those parameters of the model deemed \textit{important} when learned during the previous tasks, and forcing them to stay close to their previously computed optimal values. This penalization of changes is accomplished by adding a regularization term to the loss function while learning new tasks. The \textit{importance weight} of each parameter in the network may be determined either during or after learning a particular task. These weights are then  used to determine which parameters are more important for the preservation of the previous task(s). Competing regularization approaches differ in the specific mechanism used to compute the importance weights. Elastic weight consolidation (EWC) \cite{kirkpatrick2017overcoming}, synaptic intelligence \cite{zenke2017continual}, and memory-aware synapses \cite{aljundi2018memory} are examples of regularization-based approaches.

\textit{Generative replay-based approaches}, on the other hand, use a generative model to synthesize data from previous tasks that are regularly replayed with the original data samples from the current task, and the network parameters are jointly optimized over all data. Loss of prior knowledge is therefore avoided by periodic refreshing of the classifier's memory with data that represent such prior knowledge. Deep generative replay (DGR) \cite{shin2017continual}, deep generative replay with distillation \cite{vandeven2018generative,vandeven2019three}, and continual classification using generative models \cite{lavda2018continual} are examples of generative replay-based approaches.

Both regularization- and generative replay-based CL approaches work reasonably well in retaining prior knowledge, but their vulnerability to adversarial attacks has only recently started to be explored. For example, the vulnerability of EWC to perceptible backdoor attacks was investigated in \cite{9206809}, and the vulnerability of related importance based domain adaptation approaches to optimization based poisoning attacks has been discussed in our prior work \cite{umer2018adversarial,umer2019vulnerability}. However, to the best of our knowledge, the vulnerabilities and robustness of other regularization-based approaches, as well as those of the more successful and robust generative replay-based approaches \cite{vandeven2019three} have not yet been determined. Moreover, the vulnerabilities of these approaches to imperceptible attacks have also not been explored. We explore the vulnerabilities of these approaches to adversarial poisoning attacks under two realistic continual learning scenarios: domain and class incremental learning. 

\subsection{Adversarial Machine Learning}
Adversarial machine learning explores the vulnerabilities of machine learning algorithms to various attack scenarios, while developing possible defenses to such attacks. There are two major types of adversarial attacks \cite{huang2011adversarial}: 
in \textit{causative} (or \textit{poisoning}) attacks, the attacker adds strategically-chosen malicious data points into the training data \cite{umer2019vulnerability, biggio2012poisoning, umer2018adversarial} to adversely impact the future generalization capabilities of the classifier, whereas in \textit{exploratory} (or \textit{evasion}) attacks, the attacker manipulates the malicious data samples at test time specifically to evade detection \cite{biggio2013evasion, goodfellow2014explaining, carlini2017towards}. Since CL approaches involve retraining the model with each new batch of data (or each new task), the adversary's choice in targeting CL algorithms is typically a poisoning attack.

\textit{Backdoor attacks} are a specific class of poisoning attacks that can also be used in a hybrid poisoning-evasion scenario
\cite{gu2017badnets,shafahi2018poison}. These attacks take the form of malicious samples created by tagging a small portion of training images with a special \textit{backdoor pattern}. The adversary assigns a false label of its choice to these malicious samples, which are then added to the original training set. The model is trained on the new training dataset, which contains both correctly-labeled images as well as mislabeled, tagged images. The attacker's goal is to force the model to learn an association between the backdoor pattern and the attacker's desired class label. Once the model learns this association, it performs well on clean (untagged) test inputs during the inference stage. However, any test data that contains the attacker's chosen specific backdoor tag will be misclassified. The attacker can therefore launch \textit{targeted evasion attacks} against the model by applying the backdoor tag to any test image of its choice.


In this work, we explore the impact of the backdoor attack strategy in the context of continual learning. First, we examine the vulnerability of a variety of state of the art CL models to the gradual introduction of misinformation of backdoor samples over time. We also relax the conventional backdoor requirement of adding a large amount of mislabeled samples at once. Such a strategy allows the attacker to continuously add small amounts of misinformation over time, reducing the likelihood of detection. More importantly, unlike typical backdoor attack approaches that use visible backdoor patterns, we explore whether the models can be fooled with \textit{imperceptible backdoor patterns} in the data. Such patterns may make it more difficult to force the model to learn the misinformation, but in exchange make the  attack stealthier and more difficult to detect.

Previous works in one-shot backdoor attacks commonly assume that the attacker has access to both the training data as well as the model parameters \cite{gu2017badnets} -- an assumption that provides undue (and rather unrealistic) capabilities to the attacker. In this work, we assume the more relaxed assumption that the attacker has access only to the training data, but not to the model parameters. This more realistic setting further demonstrates the severe vulnerability of CL algorithms to backdoor attacks.

\section{Approach \& Attack Model}
We consider the problem of adversarial backdoor attacks in continual learning settings. In such settings, a new task is received through its training data $X^t$ and the corresponding labels $y^t$ at time $t$. The goal of the incremental setting is to learn a model $f_{\theta}$, parameterized by $\theta$, that minimizes the loss function 
on all tasks received at time $t$ = $1,...,T$ seen so far. The goal of the attacker is to force the model to completely forget the knowledge it acquired about a particular targeted task of attacker's own choice using imperceptible misinformation.

Conventional one-shot backdoor attacks are successful because the training and the test data are assumed to be drawn from the same distribution. In CL setting, however, there are different tasks to be learned, and the data for each task typically follows a different distribution from each other task. The attacker may select any task of its choice on which it seeks to hinder the test time performance: we refer to that task as the attacker's \textit{target task}. The attacker does not  manipulate the training data of the target task, as such obvious strategies could be too easy to detect. Rather, in our formulation, the attacker inserts malicious backdoor patterns only into the training data of the other, unrelated tasks. Hence, the attacker's challenge is to hinder the test time performance of the model on the target task using only imperceptible backdoor patterns added to the training data of tasks unrelated to the target task. 


\subsection{Attacking Regularization Based CL algorithms}
Regularization approaches find the optimal parameter vector $\theta^*$ by adding an extra regularization term to the loss function of the model. This regularization term penalizes changes to those parameters that were deemed important during the previous tasks based on the parameters' importance matrix. The model's generalized loss function $\ell(f_{\theta})$ while learning the current task at time $t$ can therefore be written as: 
\begin{equation}
\label{reg_loss}
    \ell(f_{\theta}) = \ell[f_{\theta}(X^t),y^t] + \lambda \sum_i I_{t-1,i} (\theta_{t,i} - \theta^*_{t-1,i})^2
\end{equation}
where $\ell[f_{\theta}(X^t),y^t]$ is the model's loss on the current task at time $t$;  $I_{t-1,i}$ is the $i^{th}$ parameter's importance matrix computed for the previous task at time $t-1$; $\theta^*_{t-1,i}$ is the optimal value of the $i^{th}$ parameter learned for the previous task at time $t-1$; and $\lambda$ is the regularization coefficient. We consider three popular regularization based continual learning approaches in this work: Elastic Weight Consolidation (EWC) \cite{kirkpatrick2017overcoming}, Online EWC \cite{schwarz2018progress}, and Synaptic Intelligence (SI) \cite{zenke2017continual}. It is important to mention here that each of these regularization-based approaches are based on the same principle; the main difference between them is how they compute the importance matrix. The generalized pseudo-code for regularization-based CL algorithms is provided in Algorithm \ref{Reg}.

\begin{algorithm}
\caption{Regularization-based CL algorithms}\label{Reg}
\hspace*{\algorithmicindent} \textbf{Input} $(X^t,y^t)$: Training data samples received for time step $t$; $T$: total number of tasks; $f_{\theta}$: Initial model parameterized by $\theta$; $\lambda$: regularization parameter \\
\hspace*{\algorithmicindent} \textbf{Output} Optimal Parameter $\theta^*$ and the final model $f_{\theta^*}$
\begin{algorithmic}[1]

    \For{$t= 1,...,T$:}
    \If {$t == 1$}
    \State $\theta_t^* \leftarrow \displaystyle{\minimize_{\theta} \ell[f_{\theta}(X^t),y^t]} $
    \Else
    \State Compute Fisher Information Matrix $I_{t-1}$ using optimal parameters $\theta_{t-1}^*$
    \State $\theta_t^* \leftarrow \displaystyle{\minimize_{\theta} \hspace{0.1in} \ell[f_{\theta}(X^t),y^t]} + $\par$
       \hspace{1.2in} \lambda \sum_i I_{t-1,i} (\theta_{t,i} - \theta^*_{t-1,i})^2 $
    \EndIf
    \EndFor
\end{algorithmic}
\end{algorithm}

Regularization based CL approaches are useful as they neither store data from the previous tasks nor add more layers or nodes to the network with each new incoming task, and thus avoid data storage and architectural complexity issues. However, with a fixed capacity single network and with no access to previous data at all -- not even in the pseudo form -- these approaches do struggle for challenging datasets under domain-incremental and class-incremental settings. More specifically, regularization-based approaches only perform well for those CL scenarios where the distributions between tasks are (different but) related under domain incremental learning setting and tend to fail for the more challenging class incremental learning problems -- even under no attack. We show that the attacker can take full advantage of the relatedness of data distributions among tasks, and insert a backdoor pattern with a false label into the training data of the future \textit{unrelated} tasks. These training data -- now malicious due to the backdoor patterns in them -- then serve as the misinformation. We also show that the attacker can cause significant damage by inserting as little as 1\% malicious training data, and while making the backdoor patterns imperceptible to humans.   

Mathematically, we can formally describe the attacker's objective as follows: let $X^t_{b}$ represent the malicious training data with backdoor patterns to be inserted into the training data of the current task and $y_{b}^t$ be their corresponding \textit{incorrect} labels as chosen by the attacker. The loss function to be minimized during the training of current task -- for all regularization based approaches -- is then:
\begin{equation}
\label{reg_loss_w_BD}
\begin{split}
    \ell(f_{\theta}) = &\ell[f_{\theta}(X^t \cup X^t_{b}),(y^t \cup y_{b}^t)] \\
                        & \hspace{-0.17in} + \lambda \sum_i I_{t-1,i}(\theta_{t,i} - \theta^*_{t-1,i})^2
\end{split}
\end{equation}
The pseudo-code of the generalized attack algorithm for regularization based CL approaches under this setting is shown in Algorithm \ref{Reg_att}.

\begin{algorithm}
\caption{Attacking regularization-based CL algorithms}\label{Reg_att}
\hspace*{\algorithmicindent} \textbf{Input} $(X^t,y^t)$: Training data samples received at time step $t$; $T$: total number of tasks; $f_{\theta}$: initial model parameterized by $\theta$; $\lambda$: regularization parameter \\
\hspace*{\algorithmicindent} \textbf{Output} Optimal Parameter $\theta^*$ and the final model $f_{\theta^*}$
\begin{algorithmic}[1]

    \For{$t= 1,...,T$:}
    \If {$t == 1$}
    \State $\theta_t^* \leftarrow \displaystyle{\minimize_{\theta} \ell[f_{\theta}(X^t),y^t]} $
    \Else
    \State Compute Fisher Information Matrix $I_{t-1}$ using optimal parameters $\theta_{t-1}^*$
    \State Pick 1\% random samples from the training data of the current task 
    \State Insert imperceptible backdoor pattern to these samples to create $X^t_{b}$ and assign them desired false label $y_{b}^t$
    \State Append $(X^t_{b}, y_{b}^t)$ to $(X^t, y^t)$
    \State $\theta_t^* \leftarrow \displaystyle{\minimize_{\theta}} \hspace{0.1in} \ell[f_{\theta}(X^t\cup X^t_{b}),(y^t \cup y_{b}^t)]
    $\par$ \hspace{1.05in} + \lambda \sum_i I_{t-1,i}(\theta_{t,i} - \theta^*_{t-1,i})^2 $
    \EndIf
    \EndFor
\end{algorithmic}
\end{algorithm}

\subsection{Attacking Generative Replay Based Approaches}
Generative replay-based approaches use a generator model $g$, with parameters $\phi$, to generate representative samples from previous tasks. The generated samples (from the previous tasks) are \textit{replayed}, along with the original training samples of the current task, to find the optimal parameters $\theta^*$ for the learning model $f$. We consider two common generative replay-based algorithms in this work: Deep Generative Replay (DGR) \cite{shin2017continual} and Deep Generative Replay with Distillation (DGR with Distillation) \cite{vandeven2018generative,vandeven2019three}. Both algorithms employ a variational autoencoder (VAE) \cite{kingma2013auto} to generate representative samples of previously-learned tasks. The generalized pseudo-code for generative replay-based algorithms is shown in Algorithm \ref{Gen}.

\begin{algorithm}
\caption{Generative replay-based CL algorithms}\label{Gen}
\hspace*{\algorithmicindent} \textbf{Input} $(X^t,y^t)$: Training data samples received for time step $t$; $T$: total number of tasks; $f_{\theta}$: Initial model parameterized by $\theta$; $g_{\phi}$: Initial generator parameterized by $\phi$ \\
\hspace*{\algorithmicindent} \textbf{Output} Optimal Parameter $\theta^*$ and the final model $f_{\theta^*}$
\begin{algorithmic}[1]

    \For{$t= 1,...,T$:}
    \If {$t == 1$}
    \State $\theta_t^* \leftarrow \displaystyle{\minimize_{\theta} \ell[f_{\theta}(X^t),y^t]} $
     \State $\phi_t^* \leftarrow \displaystyle{\minimize_{\phi} \ell[g_{\phi}(X^t)]} $
    \Else
    \For{k = 1,...,t-1}
    \State Generate samples from the previous task, i.e., $X^k \sim g_{\phi}$
    \State Label generated samples from the previous optimal model, i.e., $\Tilde{y}^k \leftarrow f_{\theta_{t-1}^*}(X^k)$
    \EndFor
    \State $\theta_t^* \leftarrow \displaystyle{\minimize_{\theta} \hspace{0.1in}(\frac{1}{T}\hspace{0.07in}\ell[f_{\theta}(X^t),y^t]}\hspace{0.1in} + 
    $\par$ (1-\frac{1}{T})\hspace{0.07in}\ell[f_{\theta}(X^{t-1} \cup X^{t-2} \cup ...... X^1), $\par$
    (\Tilde{y}^{t-1} \cup \Tilde{y}^{t-2} \cup ...... \Tilde{y}^1)] $)
    \State $\phi_t^* \leftarrow \displaystyle{\minimize_{\phi} \hspace{0.1in}(\frac{1}{T}\hspace{0.07in}\ell[g_{\phi}(X^t)]} + $\par$ (1-\frac{1}{T})\hspace{0.07in}\ell[g_{\phi}(X^{t-1} \cup X^{t-2} \cup ...... X^1)] $)
    \EndIf
    \EndFor
\end{algorithmic}
\end{algorithm}

Since generative replay-based approaches regularly replay (synthesized) data from previous tasks while learning the current task, they are able to regularly ``remind" the model of the prior tasks, and as such they are better able to handle the bias induced by the different, unrelated distributions between tasks. Therefore, not only can they retain (remember) the prior tasks, they also have a built-in, natural defense against an adversary that is unaware of the replay feature of the algorithm. However, most attackers are typically aware of at least the general form of the learning model. Hence, an attacker that knows that it is attacking a replay-based approach, can therefore slightly adjust its strategy to bypass the replay based built-in defense. To do so, it can simply sample and tag a small subset of training data from the target task, but still - as in attacking regularization based approaches - insert the tagged target-task images into the training data of the non-target tasks. Therefore, as the model generates and replays clean images of the target task in future tasks, the attacker will be able to continuously counter those clean images with its own tagged (malicious) images.  

In a sense, the attacker can simply hide a small number of tagged malicious images of the target task into the relatively large pool of clean (non-malicious) samples naturally generated by the algorithm's regular replay operation; and by also making the backdoor tag imperceptible, the attacker achieves a level of stealthiness that makes the attack very difficult to detect. We note that although the attacker chooses malicious samples from the training data of the target task---differently from the method for attacking regularization-based approaches---these samples are still inserted only into the  training data during training of subsequent non-target tasks. The pseudo-code for attacking generative replay-based approaches is shown in Algorithm \ref{Gen_att}.

\begin{algorithm}
\caption{Attacking generative replay-based CL algorithms}\label{Gen_att}
\hspace*{\algorithmicindent} \textbf{Input} $(X^t,y^t)$: Training data samples received for time step $t$; $T$: total number of tasks; $f_{\theta}$: Initial model parametrized by $\theta$; $g_{\phi}$: Initial generator parametrized by $\phi$ \\
\hspace*{\algorithmicindent} \textbf{Output} Optimal Parameter $\theta^*$ and the final model $f_{\theta^*}$
\begin{algorithmic}[1]

    \For{$t= 1,...,T$:}
    \If {$t == 1$}
    \State $\theta_t^* \leftarrow \displaystyle{\minimize_{\theta} \ell[f_{\theta}(X^t),y^t]} $
     \State $\phi_t^* \leftarrow \displaystyle{\minimize_{\phi} \ell[g_{\phi}(X^t)]} $
     \State Pick 1\% random samples from the training data 
    \State Insert imperceptible backdoor pattern to these samples to create $X^1_{b}$ and assign them desired false label $y_{b}^1$
    \Else
    \For{k = 1,...,t-1}
    \State Generate samples from the previous task, i.e., $X^k \sim g_{\phi}$
    \State Label generated samples from the previous optimal model, i.e., $\Tilde{y}^k \leftarrow f_{\theta_{t-1}^*}(X^k)$
    \EndFor
    \State Append $(X^1_{b}, y_{b}^1)$ to $(X^t, y^t)$ 
    \State $\theta_t^* \leftarrow \displaystyle{\minimize_{\theta}\hspace{0.1in} (\frac{1}{T}\hspace{0.07in}\ell[f_{\theta}((X^t \cup X^1_{b}),(y^t \cup y_{b}^1)]} + $\par$ (1-\frac{1}{T})\hspace{0.07in}\ell[f_{\theta}(X^{t-1} \cup X^{t-2} \cup ...... X^1), $\par$
    (\Tilde{y}^{t-1} \cup \Tilde{y}^{t-2} \cup ...... \Tilde{y}^1)] $)
    \State $\phi_t^* \leftarrow \displaystyle{\minimize_{\phi} \hspace{0.1in} (\frac{1}{T}\hspace{0.07in}\ell[g_{\phi}(X^t \cup X^1_{b})]} + $\par$ (1-\frac{1}{T})\hspace{0.07in}\ell[g_{\phi}(X^{t-1} \cup X^{t-2} \cup ...... X^1)] $)
    \EndIf
    \EndFor
\end{algorithmic}
\end{algorithm}

\section{Experiments \& Results}
To demonstrate the impact of the proposed attack, we implemented our attack strategy against several commonly-used algorithms for both regularization-based and deep generative replay-based continual learning. We use the two most common continual learning benchmark datasets for our experiments: rotation MNIST \cite{lopez2017gradient, riemer2018learning} and split MNIST \cite{zenke2017continual}. 

For rotation MNIST, we created a sequence of 5 different tasks, where each task is obtained by applying a different randomly generated rotation in the interval $(0, \pi/3]$ to the image pixels. Each task in rotation MNIST is a 10-class problem: the first task is to classify original ten MNIST digits; each subsequent task involves classifying the same ten digits with a different, but fixed, random rotation applied to all digits. 

For split MNIST, the original MNIST dataset is split into 5 different tasks, where each task is a binary classification problem: the first task is to distinguish between digits 0 and 1, the second task is to distinguish between digits 2 and 3, and so on. 

\begin{figure}
    \centering
     \begin{subfigure}[t]{0.21\textwidth}
         \centering
         \includegraphics[width=\textwidth]{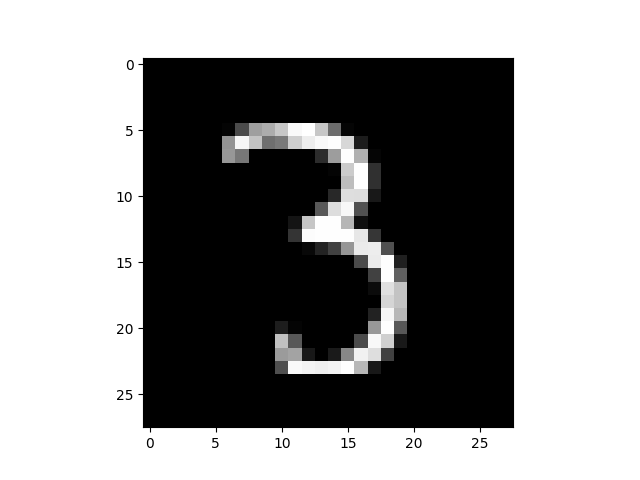}
         \caption{original image}
         \label{img_org}
     \end{subfigure}
     \hfill
     \begin{subfigure}[t]{0.21\textwidth}
         \centering
         \includegraphics[width=\textwidth]{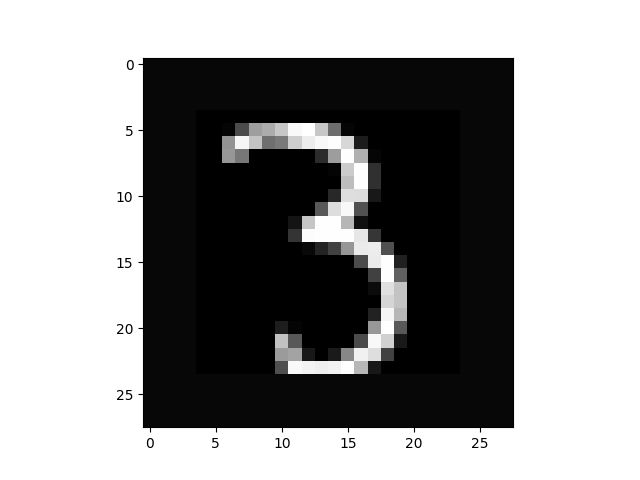}
         \caption{image with imperceptible frame as a backdoor pattern}
        \label{img_inv_fr}
     \end{subfigure}
        \caption{Imperceptible backdoor pattern for MNIST images}
        \label{fig:mnist_inv_bd}
\end{figure}

In order to insert the \textit{imperceptible backdoor pattern} into the images, we observe that most pixels around the original digit in the image are black (zero). Therefore, we use a rectangular frame around the digit with slight increase in the pixel values from 0 to 0.03. This slight increase is imperceptible to human eye \cite{fechner1948elements}, but -- as demonstrated in our results -- it is clearly visible to the classifier that directly reads pixel values. An example of an original image, and its backdoor tagged version -- with the imperceptible rectangular frame around the digit -- are shown in Figure \ref{fig:mnist_inv_bd}.

Recall that the goal of the attacker in a CL setting is to cause loss of prior knowledge, and to maximize the forgetting of a particular previous task. We refer to this prior task -- that the attacker wants the model to forget -- as the \textit{target task}. The attacker may choose any of the tasks as its desired target task. Without any loss of generality, let us assume that the target task is Task 1. The attacker could attack this target task by inserting malicious samples to the target task's training data, which will guarantee the attack to be successful (since the attack samples will come from the same distribution as the target task's test distribution). However, such an attack can be easy to detect. The attacker also wishes not to get caught, so it plans the attack to be as stealthy as possible. Therefore, in our attack model, the attacker inserts a small number of malicious examples (about 1\%) containing visually imperceptible backdoor pattern as described above into the \textit{non-target tasks} (i.e., all tasks except Task 1).
Hence, the attacker wants the model to learn the backdoor pattern and associate it with the incorrect label it had assigned while the model is learning non-target tasks. In order to determine whether the attacker has been successful in its goal, we provide backdoor tagged images, \textit{this time from the target task} to the model during test time. If the model classifies those tagged samples as the attacker's desired false label, while correctly classifying the samples of all other tasks, the attacker has been successful. Note that the attacker can also use this -- now test-time -- strategy as a means of an evasion attack, by inserting its backdoor tagged images into the target task's field (test) data. This strategy can be used, for example, to force an autonomous vehicle to recognize a stop sign with an imperceptible tag as a speed limit sign during the actual operation.



\subsection{Attacking Domain Incremental Learning}

\subsubsection{Attacking Rotation MNIST}
As described above, the attack scenario we consider allows the attacker to insert malicious samples into the training data for all tasks \textit{except} the initial target task on which the attacker wishes to inflict maximum damage at test time. In this setting, we create a small number of malicious samples by inserting an imperceptible backdoor pattern into only 1\% of the training data. Without loss of any generality, we assume that the incorrect target label that the attacker assigns to these malicious samples is of class 0. Hence, the malicious data are randomly drawn from all labels \textit{except} class 0. The falsely-labeled malicious samples (containing the imperceptible backdoor pattern) are then added to the clean training data, and the model is trained using regularization-based continual learning (CL) algorithms and generative-replay based CL algorithms. The models being attacked are therefore trained first on the clean data for the initial task, and subsequently trained on the compromised training data for each of the four subsequent tasks.

At inference time, we compute test accuracy for each task when there is ``no attack" and when there is an ``attack". Recall that the attacker's objective is to compromise Task 1; hence, in all tasks \textit{except} the first task, test examples are presented to the model \textit{without} the backdoor pattern, in order to monitor the model's ability to retain performance on untargeted tasks. For the targeted task, 100\% of the test examples from each class -- except class 0 -- are presented to the model with imperceptible backdoor tagged images. The attacker's goal here is to have each of these tagged images to be misclassified as class 0. In other words, the attacker takes control of the classifier, by forcing it to classify any pattern it desires (by tagging it with an imperceptible backdoor image) to be misclassified as any class it wishes (class 0, in this example). The test accuracy for each task, as well as the average accuracy across all tasks, under no attack and with attack are presented in Table \ref{table:accuracy_reg_rM} for regularization-based CL approaches and in Table \ref{table:accuracy_gen_rM} for generative replay-based approaches. All attack results are presented with mean and standard deviation computed over ten independent runs.

\begin{table*}[h]
\caption[]{Test accuracy on rotation MNIST using regularization-based CL approaches under domain incremental learning. The training data of every task, except the first, contains backdoor tagged images, at a ratio of 1\%.}
\label{table:accuracy_reg_rM}
\centering
\resizebox{\textwidth}{!}{\begin{tabular}{|P{0.7in}|P{0.7in}|P{0.9in}|P{0.7in}|P{0.9in}|P{0.7in}|P{0.9in}|}
\hline
Tasks & EWC under no attack & EWC under attack on Task 1 & SI under no attack & SI under attack on Task 1 & Online EWC under no attack & Online EWC under attack on Task 1 
\\ \hline
 Task 1 &  0.8817 & \textbf{0.0967 $\pm$ 0.0004}   &  0.8832 &  \textbf{0.1111 $\pm$ 0.0035} &   0.8708 & \textbf{0.0966 $\pm$ 0.0004}  \\ \hline 
Task 2 &  0.9637 & 0.9627 $\pm$ 0.0018   &  0.9513 &  0.9528 $\pm$ 0.0072 &   0.9508 &  0.9608 $\pm$ 0.0032\\ \hline
Task 3 &  0.9702 & 0.9700 $\pm$ 0.0009  &  0.9525 &  0.9522 $\pm$ 0.0036 &  0.9589 & 0.9608 $\pm$ 0.0024 \\ \hline
Task 4 &  0.9674 & 0.9679 $\pm$ 0.0007  &  0.9431  & 0.9403 $\pm$ 0.0041  &   0.9575 &  0.9529 $\pm$ 0.0025 \\ \hline
Task 5 &  0.9591 & 0.9596 $\pm$ 0.0009   &   0.9165 &  0.9084 $\pm$ 0.0114 &   0.9421 &  0.9315 $\pm$ 0.0032 \\ \hline \hline 
Avg. Accuracy &  0.9484 & 0.7917 $\pm$ 0.0009  &   0.9293 &  0.7729 $\pm$ 0.0059 &  0.9360 &  0.7805 $\pm$ 0.0023\\ \hline
\end{tabular}}
\end{table*}

\begin{table*}[h]
\caption[]{Test accuracy on rotation MNIST using generative replay based approaches under domain incremental learning. The training data of every task, except the first, contains backdoor tagged images, at a ratio of 1\%.}
\label{table:accuracy_gen_rM}
\centering
\resizebox{0.9\textwidth}{!}{\begin{tabular}{|P{0.7in}|P{0.9in}|P{0.9in}|P{0.9in}|P{0.9in}|}
\hline
Tasks & DGR under no attack & DGR under attack on Task 1 & DGR with distillation under no attack & DGR with distillation under attack on Task 1 
\\ \hline
 Task 1 & 0.9553 & \textbf{0.1161 $\pm$ 0.0074}   &  0.9683 & \textbf{0.1052 $\pm$ 0.0017}   \\ \hline 
Task 2 &  0.9728 & 0.9738 $\pm$ 0.0010   &  0.9823 &  0.9828 $\pm$ 0.0004 \\ \hline
Task 3 &  0.9742 & 0.9754 $\pm$ 0.0010   & 0.9837 &  0.9836 $\pm$ 0.0005  \\ \hline
Task 4 &  0.9753 & 0.9736 $\pm$ 0.0010   &  0.9815  &  0.9823 $\pm$ 0.0005  \\ \hline
Task 5 &  0.9688 & 0.9676 $\pm$ 0.0015   &   0.9784 & 0.9795 $\pm$ 0.0006  \\ \hline \hline
Avg. Accuracy &  0.9693 & 0.8013 $\pm$ 0.0024  &  0.9788 &  0.8067 $\pm$ 0.0007 \\ \hline
\end{tabular}}
\end{table*}

We observe from Tables \ref{table:accuracy_reg_rM} and  \ref{table:accuracy_gen_rM} that there is a dramatic drop in Task 1 performance for all five algorithms under consideration -- from upper 80\% to around 10\% for regularization approaches and from around 95\% to 10\% for replay approaches. The 10\% figure comes from the fact that all classes 1 through 9 (the classes that had the backdoor tag) are incorrectly classified (as class 0), and only the original class 0 images (which were not tagged) are classified correctly. Hence, we see that the attacker can successfully force the model to forget a specifically targeted information (Task 1, in this experiment) by corrupting only 1\% of the training data of \textit{other tasks} and further force the model to incorrectly classify those images as a class of its choosing. Importantly, the corrupted images look exactly identical to clean images to a human observer---a significant obstacle to detecting the presence of such attacks. We demonstrate that adding an imperceptible perturbation to the raw image can be easily used with malicious intent. For both regularization- and generative replay-based approaches, the attacker is extremely effective in its attack , causing all of the backdoor-tagged samples of the targeted task to be misclassified as the attacker's target label. 


\subsubsection{Attacking Split MNIST} We use the same training scenarios and setup as we used for rotation MNIST, and follow the same training protocols. As before, the attacker targets Task 1 and seeks to make the model forget what it has learned for that task by introducing imperceptible misinformation through backdoor attacks during training for subsequent tasks. Again, we assume that the incorrect target label that the attacker wishes to assign to these malicious samples is class 0. Recall that each task in split MNIST for the domain incremental scenario is a binary problem. Therefore, the class distributions among different tasks of split MNIST remain fixed under this setting. However, unlike rotation MNIST, the input marginal data distributions are not only different but also unrelated, since each task in Split MNIST contains two completely different digits. Therefore, Split MNIST poses a more challenging problem for regularization based approaches to tackle even \textit{without} any attack. Generative replay-based approaches, however, can handle this scenario (when not under attack) due to their ability to regularly replay data from the previous tasks \cite{vandeven2019three}. This is why, we only present the results of attacking generative replay based approaches, and not regularization based approaches. After all, there is no point in attacking an algorithm that fails even when there is no attack.

As before, the adversary adds 1\% malicious samples containing the imperceptible backdoor pattern into the training data of every task other than the target task. The models are trained using deep generative replay (with and without distillation), first on clean data from Task 1, and then sequentially on each of the subsequent tasks containing compromised data. To compute the test-time performance on the targeted task, clean samples from class 0 and malicious samples from class 1 (that contain the imperceptible backdoor trigger) are presented to the model. For the remaining tasks, clean examples from both classes are presented to the model at the test time. We compute the test accuracy for every task, both when there is a targeted attack to the initial task and when there in no attack. The results are shown in Table \ref{table:accuracy_gen_sM} for both DGR and DGR with distillation. As before, all results are averages of ten independent runs. 

\begin{table*}[h]
\caption[]{Test accuracy on split MNIST using generative replay based approaches under Domain Incremental Learning. The training data of every task, except the first, contains backdoor tagged images, at a ratio of 1\%.}
\label{table:accuracy_gen_sM}
\centering
\resizebox{0.9\textwidth}{!}{\begin{tabular}{|P{0.7in}|P{0.9in}|P{0.9in}|P{0.9in}|P{0.9in}|}
\hline
Tasks & DGR under no attack & DGR under attack on Task 1 & DGR with Distillation under no attack & DGR with Distillation under attack on Task 1  
\\ \hline
 Task 1 & 0.9608 & \textbf{0.4486 $\pm$ 0.0080}   &  0.9858 & \textbf{0.4537 $\pm$ 0.0033}   \\ \hline
Task 2 & 0.9540 & 0.9497 $\pm$ 0.0104   &  0.9755 &  0.9743 $\pm$ 0.0035 \\ \hline
Task 3 &  0.8895 & 0.9157 $\pm$ 0.0237   &  0.9509 &  0.9540 $\pm$ 0.0095  \\ \hline
Task 4 &  0.9844 & 0.9852 $\pm$ 0.0017   & 0.9824  & 0.9857 $\pm$ 0.0026 \\ \hline
Task 5 &  0.9788 & 0.9777 $\pm$ 0.0046   &  0.9798 &  0.9764 $\pm$ 0.0038  \\ \hline \hline
Avg. Accuracy & 0.9535 & 0.8554 $\pm$ 0.0097  &  0.9749 & 0.8688 $\pm$ 0.0045 \\ \hline
\end{tabular}}
\end{table*}

As in  the rotated MNIST example, the attack is able to successfully force the model to forget specific targeted examples from the first task, while leaving all other tasks unaffected. Not affecting other (non-target) tasks is important, as it limits the detectability of the attack \cite{detectability}. We demonstrate that because the attacker has the ability to selectively impair the model, it is able to assume complete control over the continual learner. The attacker forces all  examples from the target task containing our imperceptible trigger, to be falsely classified as the attacker's desired label. Both DGR and DGR with Distillation are reduced to achieving less than 50\% test accuracy -- i.e., below random guessing for a two-class problem - for the targeted task 1 of Split MNIST dataset.

\subsection{Attacking Class Incremental Learning}
\subsubsection{Attacking Rotation MNIST} We consider the same scenario as considered in attacking domain incremental learning, where the attacker inserts malicious samples containing imperceptible backdoor pattern to the training data of every task other than the targeted task. We force the model to learn the association between the imperceptible backdoor pattern and the attacker's desired target label (which is class "0" in our experiments). 

Once the imperceptible misinformation is learned by the model, we apply the same misinformation to 100\% of the test samples of the targeted task, excluding the samples from class 0. Table \ref{table:accuracy_gen_rM_CIL} presents the test-time performance obtained by both DGR and DGR with distillation, in both the ``attack'' and ``no attack'' scenarios. The attack is incredibly effective against both algorithms, as seen in the dramatic drop in Task 1 test performance from 70-80\% range to below 10\%, indicating that the attack forced all examples from class 1 to class 9 of Task 1 to be misclassified classified as class 0. The accuracy is less than the ideal 10\% as there were additional test examples of untargeted class 0, which are also misclassified by the model itself for reasons unrelated to the attack. 


\begin{table*}[h]
\renewcommand{\arraystretch}{1.1}
\caption[]{Test accuracy on Rotation MNIST using generative replay based approaches under class incremental learning. The training data of every task, except the first, contains backdoor tagged images, at a ratio of 1\%.}
\label{table:accuracy_gen_rM_CIL}
\centering
\resizebox{0.9\textwidth}{!}{\begin{tabular}{|P{0.7in}|P{0.9in}|P{0.9in}|P{0.9in}|P{0.9in}|}
\hline
Tasks &  DGR under no attack & DGR under attack on Task 1 & DGR with Distillation under no attack & DGR with Distillation under attack on Task 1 
\\ \hline
 Task 1 &  0.7293 & \textbf{0.0885 $\pm$ 0.0076}   &  0.8023 & \textbf{0.0876 $\pm$ 0.0035}   \\ \hline
Task 2 &  0.8339 & 0.8516 $\pm$ 0.0032   &  0.9358 &  0.9266 $\pm$ 0.0034 \\ \hline
Task 3 &  0.8725 & 0.8409 $\pm$ 0.0053  &  0.9268 &  0.9104 $\pm$ 0.0089  \\ \hline
Task 4 &  0.9306 & 0.9195 $\pm$ 0.0055   &  0.9424  & 0.9533 $\pm$ 0.0026 \\ \hline
Task 5 &  0.9707 & 0.9744 $\pm$ 0.0006   &  0.9738 &  0.9757 $\pm$ 0.0005  \\ \hline \hline
Avg. Accuracy &  0.8674 & 0.7350 $\pm$ 0.0044  & 0.9162 & 0.7707 $\pm$ 0.0038 \\ \hline
\end{tabular}}
\end{table*}

\subsubsection{Attacking Split MNIST} Again, we assume that the attacker inserts malicious samples to the training data of every other task except the target task itself. During training, the model learns to associate the backdoor pattern to the false label from the malicious samples. Once the backdoor pattern is learned, clean examples of class 0, and examples of class 1 with imperceptible backdoor pattern are presented to the model at test time for the first task. For every other task, only clean test examples, i.e., without any backdoor pattern are presented to the model. The results obtained with DGR and DGR with distillation are shown in Table \ref{table:accuracy_gen_sM_CIL}. Once again the attacker is very effective in its pursuit, and is able to inflict considerable damage on both generative replay based algorithms. Specifically, the attacker achieves its intended targeted misclassification on all class 1 examples of the initial task, hence achieving less than 50\% accuracy for task 1 at test time. The test performance is under 50\% as some of the untargeted class 0 instances are also misclassified by the model - for reasons unrelated to the attack. 

\begin{table*}[h]
\renewcommand{\arraystretch}{1.1}
\caption[]{Test accuracy on Split MNIST using generative replay based approaches under class incremental learning. The training data of every task, except the first, contains backdoor tagged images, at a ratio of 1\%.}
\label{table:accuracy_gen_sM_CIL}
\centering
\resizebox{0.9\textwidth}{!}{\begin{tabular}{|P{0.7in}|P{0.9in}|P{0.9in}|P{0.9in}|P{0.9in}|}
\hline
Tasks &  DGR under no attack & DGR under attack on Task 1 & DGR with Distillation under no attack & DGR with Distillation under attack on Task 1 
\\ \hline
 Task 1 &  0.8407 & \textbf{0.4198 $\pm$ 0.0149}   &   0.9504 & \textbf{0.4302 $\pm$ 0.0178}   \\ \hline
Task 2 &  0.8741 & 0.8692 $\pm$ 0.0241   &  0.8972 &  0.8795 $\pm$ 0.0127 \\ \hline
Task 3 &  0.8831 & 0.8529 $\pm$ 0.0383  & 0.8879 &  0.8727 $\pm$ 0.0209  \\ \hline
Task 4 &  0.9542 & 0.9394 $\pm$ 0.0140   &  0.9592  & 0.9591 $\pm$ 0.0058 \\ \hline
Task 5 &  0.9692 & 0.9749 $\pm$ 0.0058   &   0.9612 &  0.9756 $\pm$ 0.0036  \\ \hline \hline
Avg. Accuracy &  0.9043 & 0.8112 $\pm$ 0.0194  &  0.9312 & 0.8234 $\pm$ 0.0122 \\ \hline
\end{tabular}}
\end{table*}

\section{Conclusions \& Future Work}
We have shown that both regularization-based and generative replay-based continual learning approaches are vulnerable to a backdoor poisoning attack, even when only 1\% of the training data are poisoned, and perhaps more importantly, even when the backdoor patterns are completely imperceptible to humans. We demonstrate this vulnerability on two important scenarios of continual learning--- domain incremental learning and class incremental learning---using two common CL benchmark datasets. We show that an informed adversary, knowing the specific mechanisms of success of both regularization- and generative replay-based approaches, can easily exploit these mechanisms against the algorithms themselves to induce false learning with pin-point targeted damage, and force the model to forget any task and misclassify the instances of any class as any other class of its own choosing. Moreover, the imperceptible nature of the backdoor misinformation severely reduces the detectability of the attack while remaining highly damaging to the learners. Hence, this work emphasizes the critical need for continual learning algorithms seeking to be cognizant of imminent adversarial threats.

Our future work consists of exploring the vulnerabilities of continual learning approaches for other natural and realistic continual learning datasets. Perhaps more importantly, we wish to develop appropriate defensive solutions against such catastrophic attacks. 
This work further demonstrates that it is important to prioritize the development of robust learning systems, as well as defenses against adversarial attacks, such that a continual learner will be able to learn from streaming data while remaining secure against deliberate misdirections.



\bibliographystyle{IEEEtran}
%

\bibliography{bibliography}{}

\end{document}